\documentclass[letterpaper]{article}
\usepackage[preprint]{aaai2027}
\usepackage[hyphens]{url}
\usepackage{graphicx}
\usepackage{natbib}
\usepackage{caption}
\usepackage{algorithm}
\usepackage{algorithmic}
\usepackage[table]{xcolor}

\usepackage{newfloat}
\usepackage{listings}
\DeclareCaptionStyle{ruled}{labelfont=normalfont,labelsep=colon,strut=off}
\floatstyle{ruled}
\newfloat{listing}{tb}{lst}{}
\floatname{listing}{Listing}

\usepackage{booktabs}
\usepackage{multirow}
\usepackage{amsmath,amssymb}

\newcommand{\cossim}{\operatorname{sim}}

\DeclareMathOperator*{\argmax}{arg\,max}

\newcommand{\Open}{\textsc{Open}}
\newcommand{\Grounded}{\textsc{Grounded}}
\newcommand{\Unsupported}{\textsc{Unsupported}}
\newcommand{\Conflict}{\textsc{Conflict}}
\newcommand{\Unused}{\textsc{Unused}}
\newcommand{\Missing}{\textsc{Missing}}

\newcommand{\ourscell}{\cellcolor{gray!18}}

\title{Where Search Agents Go Off the Rails: Graph-Grounded Step-Level Verification}

\author{
Hyeseo Jeon\equalcontrib\textsuperscript{\rm 1},
Seonho Yoo\equalcontrib\textsuperscript{\rm 2},
Jay-Yoon Lee\corresponding\textsuperscript{\rm 1}
}
\affiliations{
\textsuperscript{\rm 1}Seoul National University\\
\textsuperscript{\rm 2}Korea University\\
\{doupari,lee.jayyoon\}@snu.ac.kr,
leoyoo2004@korea.ac.kr
}

\begin{document}

\maketitle

\begin{abstract}
Search-augmented reasoning systems suffer from unsupported intermediate claims and evidence-use failures, even when yielding correct final answers. Reliable step-level auditing requires evaluating three distinct axes: whether the evidence $D$ covers the question $Q$ ($Q \leftrightarrow D$), whether the reasoning $T$ addresses the question ($Q \leftrightarrow T$), and whether the reasoning is factually grounded in the evidence ($D \leftrightarrow T$). However, Existing methods typically assess either the intrinsic validity of reasoning steps or their support against a given evidence context, without jointly diagnosing question requirements, temporally available evidence, and reasoning claims over an evolving search trajectory. Although these verifiers often achieve high precision when predicting an error, they exhibit substantially lower recall and thus leave many erroneous trajectories undetected. This is particularly problematic for step-level auditing, where a missed error can propagate to subsequent retrieval and reasoning.

We introduce \textbf{GraphRail}, a generative-judge-free framework that decomposes each question into structured requirement triples and represents retrieved evidence and reasoning claims in the same claim form across the $Q$--$D$--$T$ relations. GraphRail diagnoses missing or unused evidence, unresolved requirements, and grounded, unsupported, or conflicting claims without using a gold answer or asking another LLM to generate a verification judgment. Across three multi-hop QA benchmarks and three search-augmented reasoning frameworks, GraphRail achieves the strongest Recall, $F_1$, $F_6$ scores in all evaluated offline settings, substantially outperforming the generative verifier baselines in recall. Beyond offline auditing, the same diagnoses can be converted into structural feedback that improves online reasoning in several settings. These results demonstrate that restricting LLMs to structured extraction and verifying with explicit cross-graph comparisons provides an effective and traceable alternative to open-ended generative judgment.
\end{abstract}

\section{Introduction}
\begin{figure}[t!]
\centering
\includegraphics[width=0.99\columnwidth]{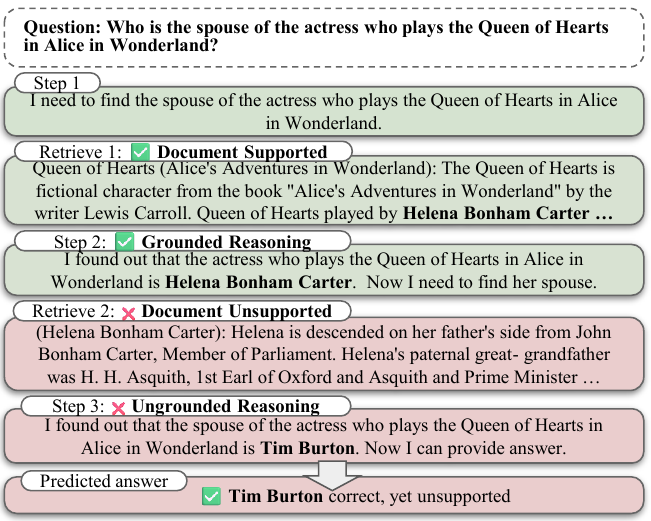}
\caption{An unsupported intermediate claim can influence subsequent retrieval and reasoning. Final-answer evaluation alone does not reveal when the trajectory loses evidence.}
\label{fig1}
\end{figure}

Search-augmented reasoning systems repeatedly alternate between reasoning and retrieval before producing a final answer. Frameworks such as Search-R1 \shortcite{jin2025search} and ReSearch \shortcite{chen2026learning} allow large language models (LLMs) to generate search queries, inspect retrieved documents, and continue reasoning over multiple turns. At any turn, the trajectory may be in one of following states: the required evidence may not yet have been retrieved, the evidence may be available but unused in reasoning, or the model may have asserted an unsupported claim. These states call for different interventions—additional retrieval, reconsideration of the available evidence, or revision of the unsupported claim—yet current systems rarely distinguish among them.
Evaluation of these states is obscured by final-answer metrics such as Exact Match and F1. Such metrics indicate whether an answer matches a reference, but not whether the evolving reasoning remains relevant to the question and supported by the retrieved context. Furthermore, a correct final answer may still be reached through an ungrounded shortcut such as Figure~\ref{fig1}. Since an early unsupported claim can influence subsequent searches and conclusions, step-level verification is therefore important for auditing whether a trajectory remains aligned with the question requirements and grounded in the evidence available at each step.

\begin{figure*}[t]
\centering
\includegraphics[width=0.99\textwidth]{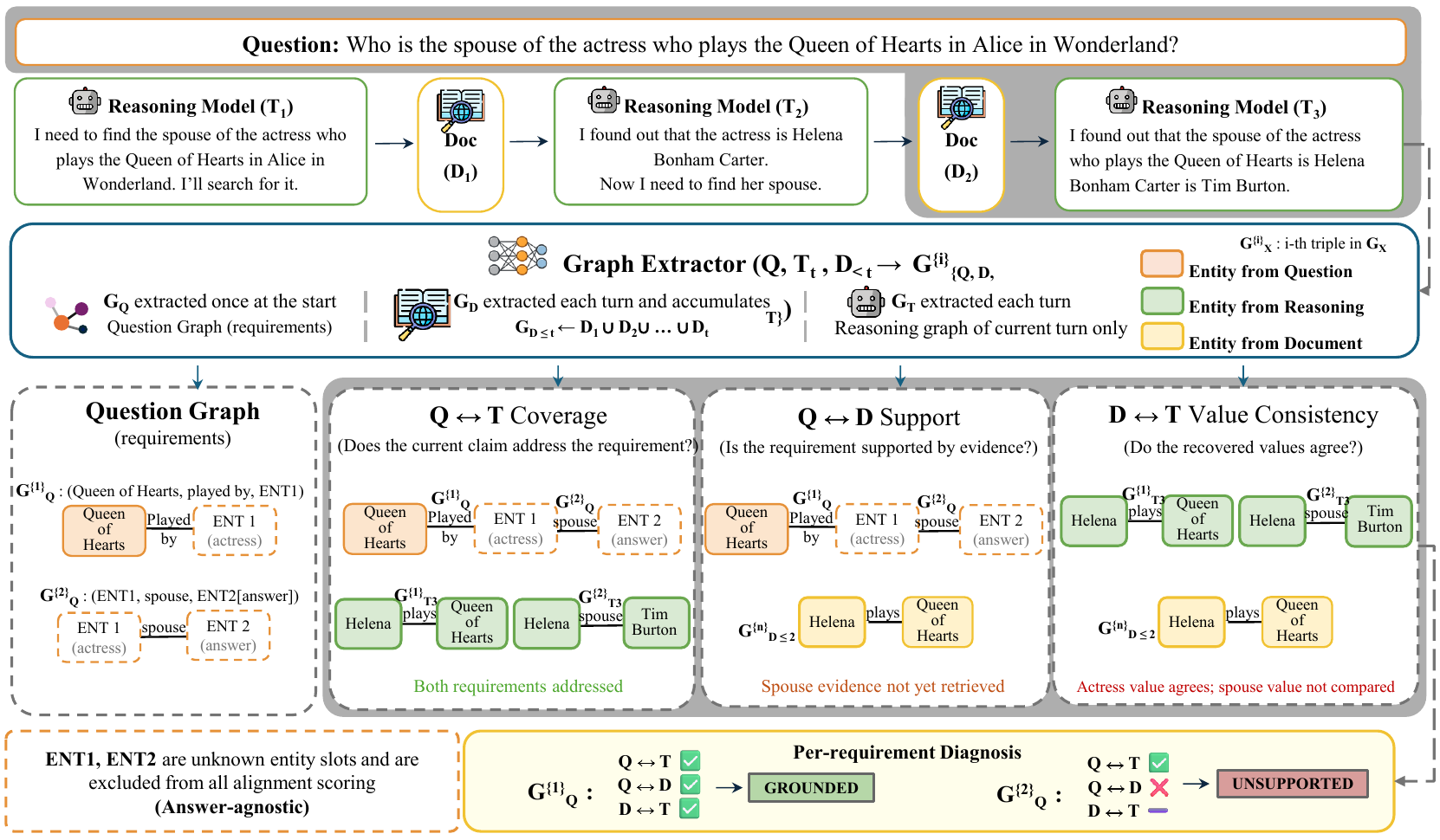}
\caption{Overview of GraphRail. Question requirements, retrieved evidence, and reasoning claims are compared through the three complementary $Q$--$D$--$T$ graph relations, enabling answer-agnostic diagnosis of 
evidence support, reasoning coverage, and value consistency without relying on an LLM-as-a-judge.}
\label{fig:graphrail}
\end{figure*}

Thus, reliable step-level verification requires jointly considering three components: the question, the retrieved documents, and the model's reasoning. We represent these components as $Q$, $D$, and $T$ respectively, and verify three relations: $Q \leftrightarrow D$, $Q \leftrightarrow T$, and $T \leftrightarrow D$. The $Q \leftrightarrow D$ relation checks whether the retrieved documents contain the evidence required by the question. The $Q \leftrightarrow T$ relation checks whether the reasoning addresses that requirement. The $T \leftrightarrow D$ relation checks whether the claims made in the reasoning are supported by the retrieved evidence.

To the best of our knowledge, existing methods verify only part of this structure by focusing on step-level reasoning, evidence grounding, or collapsing the three components into a single generative judgment. To address these limitations, we introduce \textbf{GraphRail}, a requirement-centered and temporally grounded auditing graph framework for step-level verification. GraphRail converts the question, retrieved documents, and reasoning statements into structured triple graphs. The question graph encodes the information required by the query, leaving answer entities as unknown entities; the document graph extracts facts in the retrieved evidence; and the reasoning graph extracts claims by the model. GraphRail then applies an explicit coarse-to-fine alignment procedure across the three $Q$--$D$--$T$ relations. For each question requirement, GraphRail combines the three pairwise signals into an actionable diagnosis: evidence may be \Missing{} or \Unused{}, an unresolved requirement may remain \Open{}, and a concrete claim may be \Grounded{}, \Unsupported{}, or in \Conflict{} with retrieved evidence. This alignment procedure is answer-independent: unknown answer slots are excluded from alignment, and no reference answer is used to select or verify a candidate. The same diagnosis supports both offline and online use. Offline auditing localizes where completed trajectories lose evidentiary support, while online verification provides structural feedback before the agent's next reasoning step.

Most importantly, GraphRail does not use LLM as a \textit{judge}. The role of the LLM in GraphRail is restricted to constrained graph extraction. It is not asked to decide whether another model is correct or to generate an open-ended verification judgment. This design confines model uncertainty to a bounded information-extraction stage and makes the subsequent diagnosis traceable to explicit cross-graph comparisons. As a result, GraphRail provides verification diagnosis without introducing a second generative reasoning loop or relying on the verifier’s parametric knowledge.

\smallskip
Our main contributions are as follows:
\begin{itemize}
    \item We formulate verification of evolving search trajectories as a requirement-centered diagnosis over question requirements, temporally available evidence, and current reasoning claims, separating missing or unused evidence from grounded, unsupported, and conflicting claims.
    \item We propose \textbf{GraphRail}, a gold-answer-free verification framework that confines the LLM for structured graph extraction and performs diagnosis through explicit cross-graph alignment. It can verify existing search-augmented reasoning models without modifying their parameters or architecture.
    \item We present a failure-mode analysis across three multi-hop question-answering benchmarks and evaluate GraphRail as both an offline trajectory auditor and an online source of targeted structural feedback.
\end{itemize}

\section{Related Works}

\paragraph{Process and Evidence Verification.}
\textbf{Process verification} evaluates intermediate reasoning rather than only final outputs. Foundational process reward models, developed and evaluated largely on mathematical reasoning, assign correctness scores to individual solution steps for verification or training \cite{lightman2024let,wang2024math}. Process-reward-guided search uses such signals to select promising reasoning trajectories \cite{zhang2024rest}, while graph-based verifiers model dependencies among steps or alternative reasoning paths \cite{cao2024graphreason,fang2026graph}. Socratic-PRMBench \shortcite{li2025socratic} further evaluates whether process verifiers detect errors across systematic reasoning patterns. Complementary work examines whether a final prediction actually depends on the stated reasoning chain \cite{paul2024making}. These approaches primarily evaluate the internal correctness, structure, or causal use of reasoning, rather than whether a search agent's verbal claims are supported by the evidence available when they are produced.

\textbf{Evidence-grounding} methods instead connect generated content to external sources. Prior work evaluates citation support at the answer level \cite{gao2023enabling}, introduces self-reflection over retrieved evidence at the generation-segment level \cite{asai2024self}, or penalizes unsupported reasoning steps against a provided evidence set \cite{gui2026faithrl}. Graph-based approaches also verify complex claims through multi-hop evidence \cite{jeon2025graphcheck} or interleave knowledge-graph retrieval with stepwise verification and plan revision \cite{zhao2026vog}. While these works establish increasingly fine-grained grounding, they generally verify a step, segment, or claim against a given evidence context. GraphRail instead audits
an evolving search trajectory by jointly comparing question requirements, temporally available retrieved evidence, and current reasoning claims. Its $Q$--$D$--$T$ diagnosis distinguishes missing or unused evidence from open, grounded, unsupported, and conflicting reasoning rather than reducing
verification to a single correctness or support judgment.

\paragraph{Search-Process Control and External Feedback.}
Prior studies show that intrinsic self-correction can be unreliable or even degrade reasoning performance \cite{valmeekam2023can,stechly2023gpt,huang2024large}. External-feedback methods therefore use verbal reflection, iterative refinement, tool-assisted critique, or process rewards to revise model outputs and reasoning paths \cite{shinn2023reflexion,madaan2023self,gou2024critic,zhang2024rest}. Search-specific methods further learn when to stop retrieval \cite{park2025stop}, supervise and refine intermediate search queries \cite{wen2026smartsearch}, or provide hop-aware benchmarks for diagnosing retrieval--reasoning failures \cite{you2026agenticragtracer}. These approaches primarily target correction, policy control, or benchmark-specific analysis. GraphRail instead converts explicit $Q$--$D$--$T$ diagnoses into predefined feedback templates, enabling online intervention without retraining the
underlying agent or asking another LLM to generate an open-ended critique.

\begin{table}[!t]
\centering
\setlength{\tabcolsep}{4pt}
\begin{tabular}{lccc}
\toprule
 & Hotpot & 2Wiki & Musique \\
\midrule
No Error & 45.2 & 17.4 & 6.2 \\
\specialrule{0.3pt}{2pt}{2pt}
\multicolumn{4}{c}{\textit{Error Cases}} \\
\specialrule{0.3pt}{2pt}{2pt}
Incorrect Reasoning & & & \\
\quad Given gold document       & 6.8  & 7.0  & 9.4 \\
\quad Given irrelevant document & 14.8 & 38.4 & 33.8 \\
\quad Given no evidence          & 13.0 & 14.8 & 21.6 \\
Search Query Failure             & 3.0  & 12.6 & 10.6 \\
Hop Decomposition Failure        & 5.0  & 1.8  & 10.0 \\
Decoding Failure                 & 3.0  & 2.4  & 1.8 \\
Others                           & 9.2  & 5.6  & 6.6 \\
\midrule
Accuracy (EM) & 56.2  & 38.8  & 11.2  \\
Average Model Steps & 2.57  & 3.53  & 3.58  \\
\bottomrule
\end{tabular}
\caption{Reasoning analysis for all samples. Most step-level errors arise from document hallucinations and become more frequent as average model steps increase.}
\label{tab:failure_all}
\medskip
\begin{tabular}{lccc}
\toprule
 & Hotpot & 2Wiki & Musique \\
\midrule
No Error & 80.4  & 44.9  & 56.3  \\
\specialrule{0.3pt}{2pt}{2pt}
\multicolumn{4}{c}{\textit{Error Cases}} \\
\specialrule{0.3pt}{2pt}{2pt}
Incorrect Reasoning & & & \\
\quad Given gold document & 2.5 & 3.6 & 0.0 \\
\quad Given irrelevant document & 4.3 & 29.9 & 9.1 \\
\quad Given no evidence & 11.3 & 21.6 & 31.0 \\
Search Query Failure & 0.4 & 0.0 & 0.0 \\
Hop Decomposition Failure & 0.7 & 0.0 & 3.6 \\
Decoding Failure & 0.4 & 0.0 & 0.0 \\
\bottomrule
\end{tabular}
\caption{Step-level reasoning analysis among samples with correct final answers. EM does not imply grounded reasoning}
\label{tab:failure_correct}
\end{table}

\begin{table*}[!t]
\centering
\setlength{\tabcolsep}{5pt}
\begin{tabular}{lll*{12}{c}}
\toprule
\textbf{} & \textbf{} & \textbf{} & \multicolumn{4}{c}{HotpotQA} & \multicolumn{4}{c}{2WikiMultihopQA} & \multicolumn{4}{c}{MuSiQue} \\
\cmidrule(lr){4-7} \cmidrule(lr){8-11} \cmidrule(lr){12-15}
 & & & P & R & $F_1$ & $F_6$ & P & R & $F_1$ & $F_6$ & P & R & $F_1$ & $F_6$ \\
\midrule
\multirow{6}{*}{Search-R1} & \ourscell & \ourscell 0.5B & \ourscell 56.0 & \ourscell 90.1 & \ourscell 69.1 & \ourscell 88.6 & \ourscell 86.3 & \ourscell \underline{90.1} & \ourscell 88.2 & \ourscell \underline{90.0} & \ourscell 94.4 & \ourscell \underline{97.2} & \ourscell \underline{95.8} & \ourscell \underline{97.1} \\
 & \ourscell Ours & \ourscell 4B & \ourscell 58.1 & \ourscell \textbf{96.0} & \ourscell \textbf{72.4} & \ourscell \textbf{94.3} & \ourscell 87.4 & \ourscell \textbf{90.8}& \ourscell \underline{89.1} & \ourscell \textbf{90.7} & \ourscell 94.3 & \ourscell \textbf{98.5} & \ourscell \textbf{96.4} & \ourscell \textbf{98.4} \\
 & \ourscell & \ourscell 7B & \ourscell 57.7 & \ourscell \underline{95.3} & \ourscell \underline{71.9} & \ourscell \underline{93.7} & \ourscell 87.1 & \ourscell \underline{90.1} & \ourscell 88.6 & \ourscell \underline{90.0} & \ourscell 94.6 & \ourscell 96.8 & \ourscell 95.7 & \ourscell 96.7 \\
  & Self-RAG & 7B & 66.5 & 60.9 & 63.6 & 61.0 & 87.4 & 78.9 & 82.9 & 79.1 & 95.6 & 87 & 91.1 & 87.2 \\
  & IBM-Granite & 8B & \underline{73.9} & 63.1 & 68.1 & 63.4 & \underline{92.3} & 86.7 & \textbf{89.4} & 86.8 & \underline{97.1} & 79.7 & 87.5 & 80.1 \\
  & GPT-4.1-mini & -- & \textbf{82.8} & 56.2 & 67.0 & 56.7 & \textbf{97.0} & 62.2 & 75.8 & 62.8 & \textbf{98.0} & 61.2 & 75.3 & 61.8 \\
\midrule
\multirow{6}{*}{ReSearch} & \ourscell & \ourscell 0.5B & \ourscell 48.4 & \ourscell 85.4 & \ourscell 61.8 & \ourscell 83.7 & \ourscell 80.8 & \ourscell 87.4 & \ourscell \underline{84.0} & \ourscell \underline{87.2} & \ourscell 87.1 & \ourscell 93.3 & \ourscell 90.1 & \ourscell 93.1 \\
 & \ourscell Ours & \ourscell 4B & \ourscell 48.9 & \ourscell \underline{87.1} & \ourscell \underline{62.7} & \ourscell \underline{85.3} & \ourscell 80.5 & \ourscell \underline{86.8} & \ourscell 83.6 & \ourscell 86.6 & \ourscell 87.6 & \ourscell \textbf{95.6} & \ourscell \textbf{91.4} & \ourscell \textbf{95.4} \\
 & \ourscell & \ourscell 7B & \ourscell 49.4 & \ourscell \textbf{88.0} & \ourscell \textbf{63.3} & \ourscell \textbf{86.2} & \ourscell 82.1 & \ourscell \textbf{88.7} & \ourscell \textbf{85.3} & \ourscell \textbf{88.5} & \ourscell 87.8 & \ourscell \underline{94.7} & \ourscell \underline{91.1} & \ourscell \underline{94.5} \\
  & Self-RAG & 7B & 53.2 & 74.7 & 61.7 & 73.9 & 78.0 & 84.7 & 81.2 & 84.5 & 89 & 89.4 & 89.2 & 89.4 \\
  & IBM-Granite & 8B & \underline{61.9} & 57.9 & 59.8 & 58.0 & \underline{88.1} & 77.7 & 82.6 & 77.9 & \underline{89.7} & 76.5 & 82.6 & 76.8 \\
  & GPT-4.1-mini & -- & \textbf{80.4} & 31.8 & 45.6 & 32.3 & \textbf{92.0} & 46.2 & 61.5 & 46.8 & \textbf{95.6} & 44.9 & 61.1 & 45.6 \\
\midrule
\multirow{6}{*}{ReAct} & \ourscell & \ourscell 0.5B & \ourscell 63.6 & \ourscell 87.7 & \ourscell 73.7 & \ourscell 86.8 & \ourscell 81.1 & \ourscell 85.6 & \ourscell 83.3 & \ourscell 85.5 & \ourscell 96.4 & \ourscell \underline{94.8} & \ourscell \underline{95.6} & \ourscell \underline{94.8} \\
 & \ourscell Ours & \ourscell 4B & \ourscell 65.0 & \ourscell \textbf{93.4} & \ourscell \textbf{76.6} & \ourscell \textbf{92.3} & \ourscell \underline{81.9} & \ourscell \textbf{89.1} & \ourscell \textbf{85.3} & \ourscell \textbf{88.9} & \ourscell 96.4 & \ourscell 94.6 & \ourscell 95.5 & \ourscell 94.6 \\
 & \ourscell & \ourscell 7B & \ourscell \underline{65.6} & \ourscell \underline{92.8} & \ourscell \underline{76.8} & \ourscell \underline{91.8} & \ourscell 81.6 & \ourscell \underline{86.6} & \ourscell \underline{84.0} & \ourscell \underline{86.5} & \ourscell 96.5 & \ourscell \textbf{96.7} & \ourscell \textbf{96.6} & \ourscell \textbf{96.7} \\
  & Self-RAG & 7B & 62.8 & 47.8 & 54.3 & 48.1 & 77.2 & 65.1 & 70.6 & 65.4 & 96.3 & 69.6 & 80.8 & 70.1 \\
  & IBM-Granite & 8B & \underline{65.6} & 46.2 & 54.2 & 46.6 & 80.8 & 63.0 & 70.8 & 63.4 & \underline{97.7} & 60.5 & 74.7 & 61.1 \\
  & GPT-4.1-mini & -- & \textbf{85.2} & 38.1 & 52.6 & 38.7 & \textbf{91.3} & 45.1 & 60.3 & 45.7 & \textbf{99.6} & 47.0 & 63.8 & 47.7 \\
\bottomrule
\end{tabular}
\caption{Offline GraphRail verification performance across search-augmented reasoning frameworks. P, R, $F_1$, and $F_6$ denote precision, recall, $F_\beta$ ($\beta = 1, 6$) where higher beta considers Recall with higher weight. respectively; GraphRail uses different model sizes of graph extractors: Qwen2.5-0.5B-Instruct, Qwen3-4B-Instruct-2507, and Qwen2.5-7B-Instruct, denoted as 0.5B, 4B, and 7B. We compare against generative evaluators conditioned on the evidence available before reasoning step $t$ ($Q, D_{<t}, T_t$).}
\label{tab:offline_graphrail}
\end{table*}

\section{Failure Mode Analysis}
We analyze step-level reasoning failures across HotpotQA \shortcite{yang2018hotpotqa}, 2WikiMultihopQA \shortcite{ho2020constructing}, and MuSiQue \shortcite{trivedi2022musique} using Search-R1 \shortcite{jin2025search} with a BM25 retriever.
With human evaluation of 500 samples per dataset guided by LLM-generated pseudo-labels, we group trajectories into six high-level categories. \textit{No Error} denotes valid, evidence-grounded reasoning. \textit{Incorrect Reasoning} is subdivided by the evidence source: \textit{given gold document} covers insufficient reasoning while containing gold document; \textit{given retrieved document} captures misattribution from the retrieved document; and \textit{given no evidence} covers unsupported answers from fabricated claims (hallucination). \textit{Search Query Failure} denotes repeated failed queries. \textit{Hop Decomposition Failure} includes wrong multihop question decomposition errors. \textit{Decoding Failure} captures degenerate outputs. \textit{Others} covers trajectories that are close to correct but fail during answer extraction.

Tables~\ref{tab:failure_all} and~\ref{tab:failure_correct} show that process-level errors arise from multiple
sources. As reasoning requires more steps, error-free trajectories become
rarer, while incorrect reasoning and query failures become more frequent. Even correct final answers can contain flawed intermediate
reasoning, indicating that final-answer evaluation alone is insufficient.

This motivates verifying each reasoning step with respect to three relations: question, retrieved evidence, and model's reasoning. The question--document relation checks evidence coverage, the question--reasoning relation checks whether the reasoning follows the intended requirements, and the reasoning--document relation checks whether the reasoning is grounded in evidence. Since each relation diagnoses different failures, reliable step-level verification must consider three pairwise relations.

\section{Method}
\label{sec:method}

We first introduce the graph representation used to encode the question, evidence, and reasoning. We then describe the three components of the verification framework: (1) cross-graph alignment, (2) joint requirement diagnosis, and (3) auxiliary claim grounding. Finally, we explain how this framework is applied to offline and online verification.

\begin{table*}[t]
\centering
\setlength{\tabcolsep}{2pt}
\begin{tabular}{@{}p{0.13\textwidth}cccp{0.36\textwidth}p{0.30\textwidth}@{}}
\toprule
State & $Q$--$D$ & $Q$--$T$ & $D$--$T$ & Additional Condition & Interpretation \\
\midrule
\Conflict{} & Yes & Yes & Disagree & Both aligned triples provide different concrete values for the same unknown slot. & The reasoning contradicts the available evidence. \\
\Unsupported{} & No & Yes & -- & The reasoning provides a concrete value, but no supporting document value is available. & The reasoning introduces an unsupported claim. \\
\Grounded{} & Yes & Yes & Agree / -- & No value conflict is detected; fully specified requirements do not require unknown-slot comparison. & The requirement is covered by both evidence and reasoning. \\
\Unused{} & Yes & No & -- & Supporting evidence is available, but the reasoning does not address the requirement. & Relevant evidence has been retrieved but remains unused. \\
\Open{} & No & Yes & -- & The reasoning addresses the requirement without asserting a concrete unsupported value. & The requirement remains unresolved. \\
\Missing{} & No & No & -- & Neither supporting evidence nor aligned reasoning is available. & The requirement is not covered by the current trajectory. \\
\bottomrule
\end{tabular}
\caption{Top-down requirement-level diagnosis based on $Q$--$D$ support, $Q$--$T$ coverage, and $D$--$T$ value consistency. $Q$--$D$ and $Q$--$T$ indicate whether a question requirement is aligned with a supporting document triple and a reasoning claim, respectively, while $D$--$T$ compares the concrete values recovered for the same unknown slot. ``--'' indicates that $D$--$T$ comparison is not applicable. Because some conditions overlap, the states are evaluated from top to bottom, and the first applicable state is selected.}
\label{tab:joint_diagnosis}
\end{table*}

\subsection{GraphRail}
\label{subsec:graphrail}


\subsubsection{Graph Representation.}
GraphRail represents the question, retrieved evidence, and current reasoning step as triple graphs $G_Q$, $G_D$, and $G_{T_t}$. We use $G_{D<t}$ and $G_{D\leq t}$ to denote the evidence available before and after retrieval at turn $t$. Each graph consists of triples $\tau=(h,r,o)$, where $h$, $r$, and $o$ denote the head entity, relation, and object entity respectively, separated by \texttt{[SEP]} token in sentence. Unknown answer entities in the question are represented by placeholders $\mathrm{ENT}_k$, so alignment is performed only over the known fields of each question requirement. The question graph is constructed once, the document graph accumulates evidence across turns, and the reasoning graph is constructed at each step.

\subsubsection{Verification Framework.}
\label{subsubsec:overview}

GraphRail performs requirement centered verification across the question, document, and reasoning graphs. For each question requirement $q_i \in G_Q$, the question--document ($Q$--$D$) relation checks whether relevant evidence has been retrieved, while the question--reasoning ($Q$--$T$) relation checks whether the current reasoning addresses the
requirement. When the aligned document and reasoning triples provide concrete values for the same unknown slot, the reasoning--document ($D$--$T$) relation checks whether those values agree. GraphRail combines these signals into one of six diagnostic states: \Conflict{}, \Unsupported{}, \Grounded{}, \Unused{}, \Open{}, and \Missing{}, as summarized in Table~\ref{tab:joint_diagnosis}.

\subsubsection{Cross-Graph Alignment.}
\label{subsubsec:cosine}

GraphRail uses a shared alignment operator for the $Q$--$D$ and $Q$--$T$ relations to robustly match evidence and reasoning. Given a question requirement $q_i \in G_Q$ and a candidate graph $G_C \in \{G_D, G_{T_t}\}$, the operator identifies the candidate triple that best matches the known fields of $q_i$. Unknown fields in $q_i$ are excluded from alignment.

GraphRail first linearizes each triple and retrieves the top-$K$ candidates with the highest semantic similarity to the question requirement. The shortlisted candidates are then compared at the field level. Let $q_i=(h_i,r_i,o_i)$ and $c=(h_c,r_c,o_c)$. To consider reversed subject--object orientations introduced during graph extraction, each known entity field is compared with both entity positions in $c$, while relations are compared directly:
\begin{equation}
\begin{aligned}
  s_h(q_i,c) &= \max\!\left\{\cossim(h_i,h_c),\cossim(h_i,o_c)\right\} \\
  s_o(q_i,c) &= \max\!\left\{\cossim(o_i,o_c),\cossim(o_i,h_c)\right\} \\
  s_r(q_i,c) &= \cossim(r_i,r_c)
\end{aligned}
\label{eq:cross_align}
\end{equation}

Entity fields are compared across both positions to handle reversed head--object order, while relation fields are compared directly to preserve predicate consistency. Let $\mathcal{F}(q_i,c)$ denote the fields that are known in $q_i$ and available for comparison with $c$. GraphRail scores a candidate by its weakest evaluated field. Among the top-$K$ candidates, it selects the highest-scoring candidate:

\begin{equation}
\begin{aligned}
& A(q_i,c) = \min_{f\in\mathcal{F}(q_i,c)} s_f(q_i,c) \\
& c_i^{\ast} = \argmax_{c\in\mathcal{C}_K(q_i;\mathcal{G}_C)} A(q_i,c)
\end{aligned}
\label{eq:alignment_gate}
\end{equation}

The alignment is accepted when $A(q_i,c_i^{\ast})\geq\theta$. Using the minimum field score requires every known field to match sufficiently well, preventing one strong field from masking a mismatch in another. When an accepted candidate fills an uninstantiated placeholder, the recovered value is propagated to other requirements sharing that placeholder and treated as known in their alignment. Thus, only uninstantiated placeholders are excluded, and a value recovered by the current match does not affect its alignment score. This separates requirement matching from answer-value comparison.

\subsubsection{Joint Requirement Diagnosis}
\label{subsubsec:diagnosis}

GraphRail combines $Q$--$D$ support, $Q$--$T$ coverage, and $D$--$T$ value consistency to diagnose each question requirement. For a requirement with an unknown slot, usable document support must provide a concrete value for that slot. 

The six states in Table~\ref{tab:joint_diagnosis} are evaluated in top-down priority order. As some conditions overlap, the first applicable state is selected. A $D$--$T$ value mismatch is labeled \Conflict{}, while a concrete value recovered through $Q$--$T$ without corresponding $Q$--$D$ support is labeled \Unsupported{}. If neither condition applies, the remaining states distinguish whether the requirement is covered by the document graph, the reasoning graph, both, or neither. 

The $D$--$T$ comparison applies only when the requirement contains an unknown answer slot and both aligned triples provide concrete values for that slot. Fully specified requirements are diagnosed using $Q$--$D$ support and $Q$--$T$ coverage.

\subsubsection{Auxiliary Claim Grounding}
\label{subsubsec:direct_claim}

In addition to requirement-centered diagnosis, GraphRail applies an auxiliary grounding check to additional concrete claims introduced during reasoning. Each eligible claim is compared with the available document graph using the same semantic retrieval and field-level matching components as cross-graph alignment. For conflict detection, structural compatibility and value agreement are evaluated separately: candidate document triples are first matched using the shared entity and relation fields, after which their concrete values are compared. The claim is checked against the evidence available at the corresponding auditing stage: $G_{D<t}$ offline and $G_{D\leq t}$ online. Eligible claims must be fully instantiated and explicitly asserted; triples containing unknown fields or merely restating the question are excluded. A claim is labeled \Unsupported{} when no compatible document evidence supports the asserted value, and \Conflict{} when structurally compatible evidence provides a different concrete value. This auxiliary check extends grounding verification to additional intermediate assertions.

\subsubsection{Offline GraphRail Verification}
\label{subsubsec:offline_audit}

GraphRail audits a completed trajectory using only the evidence available at the time each reasoning step was generated. At turn $t$, claims in $G_{T_t}$ are grounded against $G_{D<t}$, which contains only evidence retrieved before that reasoning step. Newly retrieved document triples are added for subsequent turns, but later evidence cannot retrospectively support an earlier claim.

The offline audit consists of two stages. First, a per-turn grounding audit detects \Unsupported{} and \Conflict{} claims as the trajectory unfolds. Second, after the final turn, a requirement-level coverage audit identifies \Unused{} requirements, for which supporting evidence was retrieved but not
used in reasoning, and \Missing{} requirements, for which the required evidence was never retrieved. 
A trajectory is marked as erroneous if any turn contains an \Unsupported{} or \Conflict{} claim, or if the final coverage audit contains a \Missing{} requirement. \Unused{} is retained as a diagnostic state but is not treated as a case-level error because the relevant evidence is already available.

\subsubsection{Online GraphRail Verification}
\label{subsubsec:online_feedback}

In the online setting, GraphRail evaluates each reasoning step against the evidence accumulated through the current turn, $G_{D\leq t}$, and returns the resulting diagnosis before the agent takes its next action. Unlike offline auditing, this setting uses newly retrieved evidence immediately to guide subsequent reasoning.

Each diagnostic state is mapped to a predefined feedback template: \Missing{} and \Open{} indicate that additional evidence is needed, \Unused{} indicates that retrieved evidence should be incorporated into the reasoning, and \Unsupported{} and \Conflict{} indicate that a concrete claim should be reconsidered. \Grounded{} requires no corrective feedback. The selected template is appended to the input containing the next retrieved document, providing verification steering guidance without directly generating or overriding the model's subsequent search and reasoning actions.

\section{Experiments}
\subsection{Experimental Setup}

\paragraph{Graph Extractor.}
We fine-tune separate question, document, and reasoning extractors as LoRA adapters over a shared base model. We evaluate Qwen2.5-0.5B-Instruct, Qwen3-4B-Instruct-2507, and Qwen2.5-7B-Instruct. The document extractor outputs factual triples from retrieved passages, the question extractor produces requirement triples with $\mathrm{ENT}_k$ placeholders, and the reasoning extractor extracts claims asserted in each reasoning step.

To construct supervision, we first collect Search-R1 trajectories and prompt Qwen2.5-7B-Instruct separately for the question, document, and reasoning extraction tasks. The resulting task-specific graphs are used as supervision
for training the corresponding extractors. For each extraction task, we build 4,500 instances from HotpotQA (1,000), 2WikiMultihopQA (1,500), and MuSiQue (2,000). We then fine-tune the extractors for three epochs with LoRA
($r=16$, $\alpha=32$, dropout $=0.1$), using a learning rate of $10^{-4}$ on a single NVIDIA RTX PRO 6000 GPU.

\paragraph{Verification.}
GraphRail embeds triples with BGE-large-en-v1.5, retains the top $K=3$ candidates, and accepts alignments with a field-level score of at least $\theta=0.60$. 

\paragraph{Evaluation Data.}
We audit Search-R1, ReSearch, and ReAct on HotpotQA, 2WikiMultihopQA, and MuSiQue using 500 vanilla trajectories per framework--dataset pair. All systems use BM25 retrieval over the Wikipedia-2018 dump. Since these benchmarks do not provide trajectory-level error labels, we manually annotate the evaluated
trajectories to construct human reference labels for offline verification. Search-R1 trajectories are additionally categorized by failure type for the failure-mode analysis.

\subsection{Results}

\begin{table}[t]
\centering
\setlength{\tabcolsep}{2pt}
\begin{tabular}{lllcccccc}
\toprule
\multirow{2}{*}{} & \multirow{2}{*}{} & \multirow{2}{*}{} & \multicolumn{2}{c}{Hotpot} & \multicolumn{2}{c}{2Wiki} & \multicolumn{2}{c}{MuSiQue} \\
\cmidrule(lr){4-5} \cmidrule(lr){6-7} \cmidrule(lr){8-9}
 & & & EM & F1 & EM & F1 & EM & F1 \\
\midrule
\multirow{4}{*}{Search-R1} & Vanilla & -- & 56.6 & 63.0 & 38.8 & 42.3 & 11.2 & 17.4 \\
 & \ourscell & \ourscell 0.5B & \ourscell \underline{57.4} & \ourscell 63.5 & \ourscell \underline{46.2} & \ourscell \underline{49.3} & \ourscell 13.4 & \ourscell \underline{20.0} \\
 & \ourscell Ours & \ourscell 4B & \ourscell \underline{57.4} & \ourscell \textbf{63.9} & \ourscell 45.8 & \ourscell 48.8 & \ourscell \textbf{14.6} & \ourscell \textbf{20.5} \\
 & \ourscell & \ourscell 7B & \ourscell \textbf{57.8} & \ourscell \underline{63.8} & \ourscell \textbf{46.4} & \ourscell \textbf{49.4} & \ourscell \underline{13.6} & \ourscell 19.7 \\
\midrule
\multirow{4}{*}{ReSearch} & Vanilla & -- & 56.4 & 64.0 & 49.4 & 52.0 & \underline{19.4} & \textbf{28.0} \\
 & \ourscell & \ourscell 0.5B & \ourscell \underline{58.0} & \ourscell \underline{64.8} & \ourscell \textbf{53.6} & \ourscell 55.7 & \ourscell \textbf{19.6} & \ourscell \textbf{28.0} \\
 & \ourscell Ours & \ourscell 4B & \ourscell \textbf{59.6} & \ourscell \textbf{66.8} & \ourscell \textbf{53.6} & \ourscell \textbf{55.8} & \ourscell 17.6 & \ourscell 26.2 \\
 & \ourscell & \ourscell 7B & \ourscell 55.6 & \ourscell 63.9 & \ourscell \underline{53.0} & \ourscell \underline{55.4} & \ourscell \underline{19.4} & \ourscell \underline{27.7} \\
\midrule
\multirow{4}{*}{ReAct} & Vanilla & -- & \textbf{42.0} & \textbf{47.9} & \textbf{42.8} & \textbf{45.2} & 7.6 & 13.0 \\
 & \ourscell & \ourscell 0.5B & \ourscell 40.2 & \ourscell 46.1 & \ourscell \underline{41.4} & \ourscell \underline{44.1} & \ourscell 9.2 & \ourscell 14.2 \\
 & \ourscell Ours & \ourscell 4B & \ourscell 39.0 & \ourscell 46.0 & \ourscell 39.8 & \ourscell 43.1 & \ourscell \underline{10.0} & \ourscell \underline{14.3} \\
 & \ourscell & \ourscell 7B & \ourscell \underline{40.4} & \ourscell \underline{46.5} & \ourscell 41.0 & \ourscell 43.6 & \ourscell \textbf{11.0} & \ourscell \textbf{14.8} \\
\bottomrule
\end{tabular}
\caption{Online GraphRail results across search-augmented reasoning frameworks. Without directly overriding in the model's reasoning process, structural feedback can improve EM/F1.}
\label{tab:online_graphrail}
\end{table}

\begin{table}[!t]
\centering
\setlength{\tabcolsep}{4pt}
\begin{tabular}{lcccc}
\toprule
Method & Hotpot & 2Wiki & MuSiQue & \shortstack{Decode\\Tokens} \\
\midrule
\rowcolor{gray!18} Ours (0.5B) & 1.18s & 1.28s & 1.19s & 1,503 \\
\rowcolor{gray!18} Ours (4B) & 3.11s & 3.43s & 3.17s & 1,531 \\
\rowcolor{gray!18} Ours (7B) & 4.13s & 4.59s & 4.18s & 1,532 \\
Self-RAG (7B) & 0.26s & 0.35s & 0.33s & 163 \\
IBM-Granite (8B) & 0.17s & 0.20s & 0.21s & 3 \\
GPT-4.1-mini & 2.28s & 2.83s & 3.14s & 53 \\
\bottomrule
\end{tabular}
\caption{Average sample offline latency and decode tokens.}
\label{tab:offline_latency}
\medskip
\setlength{\tabcolsep}{2pt}
\begin{tabular}{@{}lccc@{}}
\toprule
Method & Precision & Recall & $F_6$-Score \\
\midrule
\multicolumn{4}{c}{Temperature = 1.0} \\
\midrule
\rowcolor{gray!18} Ours (0.5B) & 79.1 ($\pm$ \underline{0.3}) & 93.4 ($\pm$ \underline{0.8}) & 92.8 ($\pm$ 0.8) \\
\rowcolor{gray!18} Ours (4B) & 79.1 ($\pm$ \textbf{0.2}) & \textbf{94.0} ($\pm$ \textbf{0.4}) & \textbf{93.4} ($\pm$ \textbf{0.4}) \\
\rowcolor{gray!18} Ours (7B) & 79.8 ($\pm$ \underline{0.3}) & \underline{93.9} ($\pm$ \textbf{0.4}) & \underline{93.3} ($\pm$ \underline{0.4}) \\
Self-RAG (7B) & \underline{83.9} ($\pm$ 1.4) & 71.3 ($\pm$ 8.1) & 71.6 ($\pm$ 7.9) \\
IBM-Granite (8B) & \textbf{88.4} ($\pm$ 1.1) & 75.7 ($\pm$ 3.3) & 75.4 ($\pm$ 3.6) \\
\midrule
\multicolumn{4}{c}{Temperature = 0.5} \\
\midrule
\rowcolor{gray!18} Ours (0.5B) & 79.1 ($\pm$ 0.4) & 92.7 ($\pm$ 0.5) & 92.2 ($\pm$ 0.5) \\
\rowcolor{gray!18} Ours (4B) & 79.4 ($\pm$ \textbf{0.1}) & \textbf{94.4} ($\pm$ \textbf{0.2}) & \textbf{93.8} ($\pm$ \textbf{0.2}) \\
\rowcolor{gray!18} Ours (7B) & 80.0 ($\pm$ \underline{0.2}) & \underline{93.7} ($\pm$ \underline{0.4}) & \underline{93.1} ($\pm$ \underline{0.4}) \\
Self-RAG (7B) & \underline{83.7} ($\pm$ 0.7) & 73.4 ($\pm$ 3.9) & 73.7 ($\pm$ 3.8) \\
IBM-Granite (8B) & \textbf{87.6} ($\pm$ 1.9) & 76.2 ($\pm$ 1.6) & 76.4 ($\pm$ 1.5) \\
\bottomrule
\end{tabular}
\caption{Uncertainty robustness of offline GraphRail verification in Search-R1, averaged over three datasets. Each entry reports mean $\pm$ standard deviation over five seeds. By using LLMs only as graph extractors, GraphRail consistently exhibits the lowest variance, indicating robustness.}
\label{tab:temperature_ablation}
\end{table}

\subsubsection{Offline GraphRail Verification}

Table~\ref{tab:offline_graphrail} evaluates trajectory level error detection against human reference labels. Recall is particularly important for offline verification because a missed error allows an unsupported, conflicting, or incomplete trajectory to remain undetected. Following \cite{song2026retrieving}, we note that multi-hop QA is particularly sensitive to missing required evidence and that recall-weighted $F_\beta$ better reflects answerability than precision alone. Because missed trajectory errors can remain undetected and affect subsequent reasoning, we report $F_6$ to place greater emphasis on recall while still penalizing false positives.

GraphRail achieves the highest $F_6$ across all framework--dataset pairs, driven by consistently strong recall. The improvement is especially pronounced on MuSiQue, where GraphRail maintains recall above 93 across all evaluated frameworks and extractor sizes. Notably, even the 0.5B extractor outperforms the larger generative verifier baselines in $F_6$ across all settings. This result supports GraphRail's design of restricting the learned model to structured graph extraction and performing verification through explicit alignment and diagnosis rules. Increasing the extractor size to 4B generally provides the strongest overall results, while the 7B extractor yields no consistent additional improvement. Notably, even the 0.5B extractor exceeds all generative verifier baselines in
$F_6$ across the nine evaluation settings, showing that strong verification
does not require a large generative judge.

\subsubsection{Online GraphRail Feedback}

Table~\ref{tab:online_graphrail} shows that GraphRail feedback consistently improves Search-R1 and benefits ReSearch on HotpotQA and 2WikiMultihopQA. For Search-R1 on 2WikiMultihopQA, EM increases from 38.8 to 46.4 and F1 from 42.3 to 49.4. ReAct, in contrast, improves only on MuSiQue
and declines on HotpotQA and 2WikiMultihopQA. This difference can be explained
by how each framework exposes and updates its intermediate state. Search-R1
and ReSearch generate factual assertions and subsequent plans within a
continuous \texttt{<think>} span followed by a free-form \texttt{<search>}
query, allowing feedback to revise both within a single generation. ReAct
instead separates an undelimited \texttt{Observation:} from a constrained
\texttt{Action: search[\ldots]} slot, causing feedback to blend with retrieved
evidence and requiring its translation into the action format. This interface
overhead limits the benefit of feedback when there is less room for correction.

\subsubsection{Offline Latency}

Table~\ref{tab:offline_latency} reports warmed-up end-to-end latency per trajectory. For all methods, latency is averaged per sample with a single active client per server. For GraphRail, question, document, and reasoning graph extraction calls are issued concurrently through the same vLLM server, and the reported latency includes the subsequent verification stage.

The higher latency relative to the local verifier baselines mainly reflects the decoding workload. GraphRail generates approximately 1,500 structured output tokens per trajectory, whereas Self-RAG, IBM-Granite, and GPT-4.1-mini generate 163, 3, and 53 tokens, respectively. Thus, GraphRail constructs explicit question, document, and reasoning graphs rather than producing only a short verification output. Nevertheless, the 0.5B variant requires only about 1.2 seconds per sample and already outperforms all evaluated baselines in offline $F_6$, providing a performance--latency balance.

\subsubsection{Offline Uncertainty Robustness}

Table~\ref{tab:temperature_ablation} evaluates the stability of offline verification under stochastic decoding at temperatures 0.5 and 1.0 over five random seeds. Across all extractor sizes, GraphRail exhibits consistently low variance in precision, recall, and $F_2$. In particular, its $F_2$ standard deviation remains at or below 0.6, while the generative verifier baselines show larger fluctuations, especially in recall.

GraphRail also maintains similar mean performance across the two temperatures. For example, the 4B extractor achieves an $F_2$ of $89.8 \pm 0.3$ at temperature 1.0 and $90.2 \pm 0.1$ at temperature 0.5. This suggests that stochastic variation in the extracted triples has limited impact on the final trajectory-level diagnosis. Since the LLM is used to produce structured graphs rather than an open-ended verification judgment, the subsequent deterministic alignment and diagnosis stages can yield stable decisions across decoding settings.






\section{Conclusion}
We presented GraphRail, a graph-grounded framework that verifies search-augmented reasoning through explicit alignment among question requirements, retrieved evidence, and reasoning claims. By combining the $Q$--$D$, $Q$--$T$, and $D$--$T$ relations, GraphRail distinguishes missing or unused evidence from grounded, unsupported, and conflicting reasoning without using a generative judge or gold answer.

GraphRail achieves strong recall and $F_6$ for offline trajectory auditing, remains stable under stochastic graph extraction, and provides useful online feedback for reasoning frameworks that expose explicit intermediate claims and plans. These findings suggest that confining the LLM to graph extraction and performing verification through deterministic cross-graph comparisons offers a traceable and effective alternative to open-ended generative verification.

\bibliography{aaai2027}

\end{document}